%% file: main.tex
\documentclass[10pt,twocolumn,letterpaper]{article}

\usepackage[pagenumbers]{cvpr} %

\definecolor{cvprblue}{rgb}{0.21,0.49,0.74}
\usepackage[pagebackref,breaklinks,colorlinks,allcolors=cvprblue]{hyperref}

\def\paperID{16873} %
\def\confName{CVPR}
\def\confYear{2025}

\usepackage{amsmath}
\usepackage{amssymb}
\usepackage[textwidth=18mm]{todonotes}
\usepackage[capitalize]{cleveref}
\usepackage{xspace}
\usepackage{colortbl}
\usepackage{subcaption}
\usepackage{bbm}
\usepackage{graphicx}
\usepackage{wrapfig}
\usepackage{enumitem}
\usepackage{booktabs}
\usepackage{listings} %
\usepackage{mathtools}
\usepackage{multirow}

\definecolor{lightgray}{gray}{0.95}
\newcolumntype{P}[1]{>{\centering\arraybackslash}m{#1}}
\newcolumntype{L}[1]{>{\raggedright\arraybackslash}m{#1}}

\title{From Multimodal LLMs to Generalist Embodied Agents: Methods and Lessons}

\author{
  Andrew Szot$\thanks{Core contributor}^{\  1,2}$ \quad Bogdan Mazoure$^{*1}$ \quad Omar Attia$^1$ \quad Aleksei Timofeev$^1$  \\ Harsh Agrawal$^1$  \quad Devon Hjelm$^1$  \quad Zhe Gan$^1$ \quad Zsolt Kira$^2$ \quad Alexander Toshev$^1$ \\
$^1$ Apple, $^2$ Georgia Tech \\
\textit{a.szot@apple.com, toshev@apple.com}
}

\newcommand{\mllm}{MLLM\xspace}
\newcommand{\mllms}{MLLMs\xspace}

\newcommand{\gea}{GEA\xspace}
\newcommand{\geabase}{GEA-Base\xspace}
\newcommand{\geabasesmall}{GEA-Base-500m\xspace}
\newcommand{\geasmallbase}{GEA-Base-500m\xspace}
\newcommand{\geafull}{Generalist Embodied Agent\xspace}

\newcommand{\lov}{LLaVA-OneVision\xspace}
\newcommand{\lovsmall}{LLaVA-OneVision-500m\xspace}

\definecolor{Gray}{gray}{0.95}
\definecolor{NoteColor}{gray}{0.4}
\definecolor{NoComp}{gray}{0.5}

\begin{document}

\maketitle

\input{sections/abstract}
\input{sections/intro}

\input{sections/related_work}

\input{sections/method}

\input{sections/environments}

\input{sections/experiments}

\input{sections/conclusion}

\input{sections/acknowledgements}
{
    \small
    \bibliographystyle{ieeenat_fullname}
    \bibliography{main}
}

\newpage
\appendix

\input{supp/method}

\input{supp/datasets}
\input{supp/experiments}

\end{document}

%% file: sections/abstract.tex
\begin{abstract}
    We examine the capability of Multimodal Large Language Models (MLLMs) to tackle diverse domains that extend beyond the traditional language and vision tasks these models are typically trained on. Specifically, our focus lies in areas such as Embodied AI, Games, UI Control, and Planning. To this end, we introduce a process of adapting an MLLM to a Generalist Embodied Agent (GEA). GEA is a single unified model capable of grounding itself across these varied domains through a multi-embodiment action tokenizer. GEA is trained with supervised learning on a large dataset of embodied experiences and with online RL in interactive simulators. We explore the data and algorithmic choices necessary to develop such a model. Our findings reveal the importance of training with cross-domain data and online RL for building generalist agents. The final GEA model achieves strong generalization performance to unseen tasks across diverse benchmarks compared to other generalist models and benchmark-specific approaches.
\end{abstract}

%% file: sections/intro.tex
\section{Introduction}\label{sec:intro}

\begin{figure}[t]
    \centering
    \includegraphics[width=0.4\textwidth]{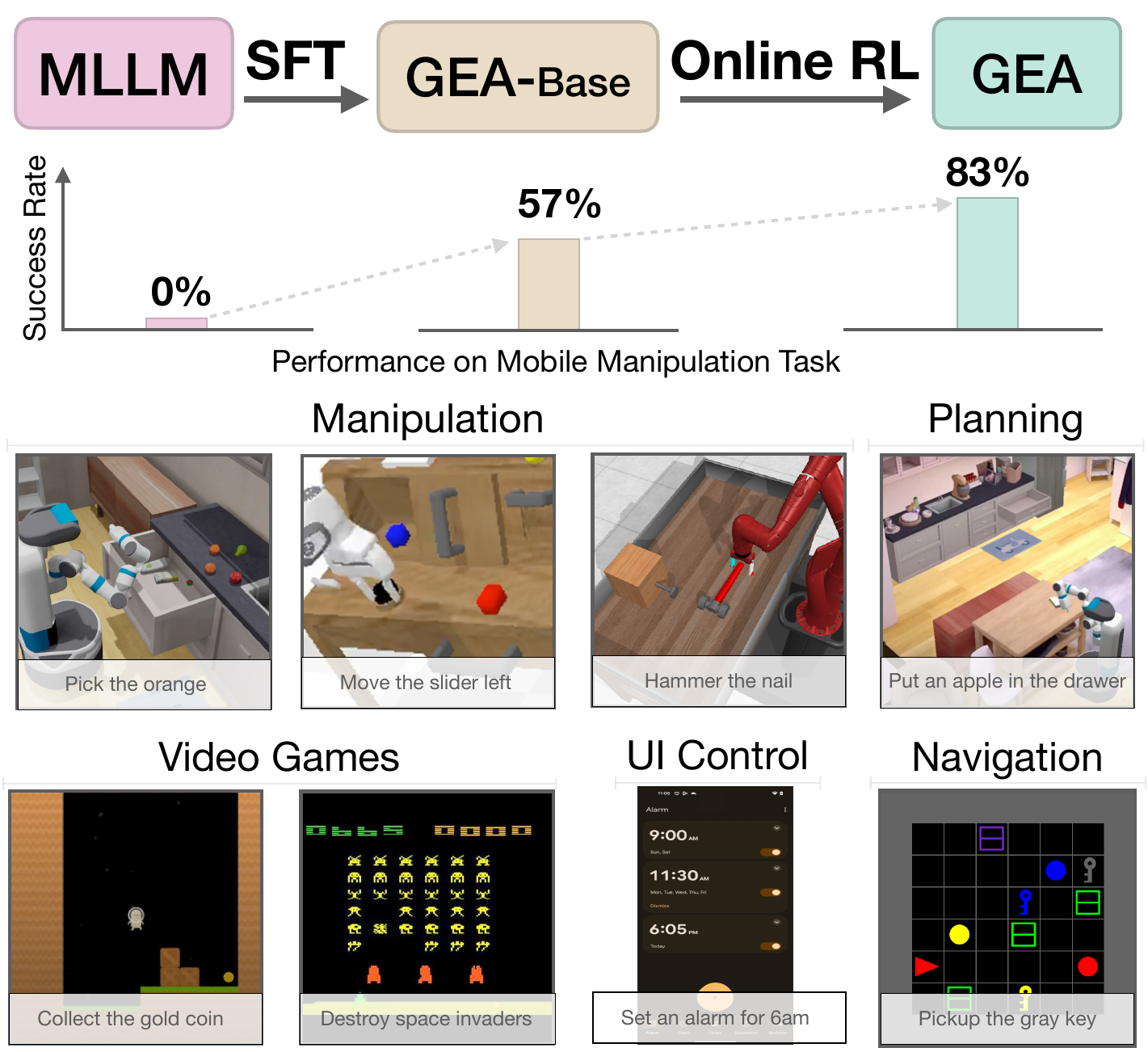}
    \caption{
        The \geafull (\gea) is a multimodal LLM-based agent that can complete tasks from natural language instructions across a variety of domains and embodiments spanning manipulation, planning, game playing, and UI control. A pretrained MLLM is finetuned with supervised finetuning (SFT) on a large dataset of embodied experiences. The final \gea model is then finetuned with reinforcement learning (RL). \gea achieves competitive results in generalization to unseen settings.
    }
    \label{fig:intro-fig}
\end{figure}

Foundation Models have demonstrated broad capabilities across language and image understanding tasks~\citep{bai2023qwen,kosmos-1,peng2023kosmos2,blip-2,instruct-blip,llava,li2023multimodal,zhu2023minigpt,ye2023mplug,li2023otter,li2023mimic, mckinzie2024mm1,idefics}. In particular, Multimodal LLMs (MLLMs) -- multimodal foundation models trained on vast amounts of textual and image data -- excel at tasks that are natural to their text and image training modalities.
As an extension of MLLMs, Vision-Language-Action models have been successfully applied in robotics and embodied AI~\citep{ahn2022can,brohan2023rt,driess2023palm,szot2023large}, as well as agents for the web ~\citep{shaw2023pixels,zheng2024gpt,koh2024visualwebarena,he2024webvoyager} and user interface (UI) control~\citep{rawles2024androidinthewild,hong2024cogagent,wen2023empowering}. 

\looseness=-1 These applications have demonstrated that MLLMs can be successfully applied to diverse domains for the purpose of controlling various embodiments like robots, playing games, and controlling devices via UIs.
As many of these domains share similarities, it is natural to ask how a single agent can be trained to be generally proficient in all of these domains.
This is a challenging problem as many of these tasks require physics and geometric reasoning, their embodiments are either static or share morphologies via a mobile manipulator, their applications require long-horizon planning, and many are partially observable and require reasoning over long sequences of observations.
In addition, training with combined data across domains with these similarities may yield cross-domain benefits, where a single agent may outperform agents trained on individual domains. 

In this work, we demonstrate an approach for adapting an MLLM into a single \textbf{G}eneralist \textbf{E}mbodied \textbf{A}gent~(\gea) to solve a vast number of tasks across diverse domains spanning manipulation, navigation, video game playing, and UI control. To enable \gea to control diverse embodiments, we learn a unified learned tokenization mechanism across all continuous and discrete action spaces. As \Cref{fig:intro-fig} illustrates, we then employ supervised finetuning (SFT)~\cite{wei2021finetuned} to adapt a pretrained MLLM to predict actions from trajectories of agents successfully completing tasks. This SFT dataset spans over 2.2 million trajectories from diverse collection methods like human labelers or learned policies. While this SFT process produces a capable agent, it suffers from an inherent lack of data, specifically data diversity, and rarely exhibits robustness to mistakes. We thus also train \gea with a second stage of online reinforcement learning (RL) training over a subset of the domains where the agent collects and learns from data in interactive simulators.

\looseness=-1 We demonstrate that GEA exhibits strong generalist capabilities. Specifically, it reaches state-of-the-art performance across many benchmarks against other generalist agents and even outperforms or closely matches bespoke specialist systems. For example, in the CALVIN manipulation benchmark~\cite{mees2022calvin}, \gea reaches $ 90 \%$ success rate while operating on unseen instructions and background, which is nearly $ 10\%$ higher than similar methods~\cite{li2023vision}  and closely matches the performance of specialist systems~\cite{ke20243d}. In a Habitat mobile pick task~\cite{szot2021habitat}, \gea achieves $ 83\%$ success in unseen scenes, outperforming a policy trained with RL on the ground truth simulator state. Similarly, in Procgen video games~\cite{cobbe2020leveraging} \gea reaches $44 \%$ of expert score, which is almost $ 20\%$ higher than prior specialist~\cite{mediratta2023generalization} models.

We also analyze the relationship between the generalist capabilities of \gea and its training data and base \mllm. We demonstrate that training on the combined data from a diverse set of domains for SFT provides a cross-domain performance boost over using only per-domain data. 
 Finally, we explore the role of RL and online data collection for building generalist agents and empirically demonstrate the benefits of online RL over prior approaches of iterative SFT or offline RL.

As a further contribution to the community, we will release the code for training and evaluating GEA along with the GEA model itself. We will add the link to the code and model to this paper when they are ready for release.

%% file: sections/related_work.tex
\section{Related Work}

Prior works have explored building generalist agents by training policies on large multi-task datasets, illustrating the importance of scaling interactive data to create capable multi-task agents~\cite{reed2022generalist,gallouedec2024jack,wei2023imitation}. Additionally, prior works have studied new architectures for generalist agents~\cite{wang2024scaling,haldar2024baku}, while others focus on applying generalist agents to robotic contexts~\cite{team2024octo,brohan2022rt,bousmalis2023robocat}. Some research also investigates generalist models in domain-specific benchmarks~\cite{jiang2022vima,team2023human} or in cross-embodiment scenarios~\cite{patel2024get,doshi2024scaling}. Like \gea, these approaches leverage extensive data, yet our work emphasizes the importance of adapting a pretrained \mllm via both finetuning and online RL. 
For example, differences between \gea and Gato~\cite{reed2022generalist}, are that \gea leverages RL, utilizes a pretrained MLLM, learns a multi-embodiment action tokenizer, and focuses on evaluating generalization to new task settings. As a result, \gea empirically outperforms Gato in many settings. 

Like \gea, some prior work focuses on adapting \mllms as agents. Works have proposed domain-specific pipelines for using the zero-shot capabilities of \mllms for decision-making~\cite{huang2022language,ahn2022can,zeng2022socratic,liang2023code,wu2023tidybot,wang2023voyager}, while our work focuses on finetuning \mllms. Other works investigate schemes for finetuning \mllms for decision-making and the benefits of doing so, but also in the context of specific domains~\cite{li2023vision,shi2023unleashing,szot2024grounding,szot2023large}. Prior work also finetunes \mllms as generalist agents~\cite{kim2024openvla,brohan2023rt,o2023open}. However, our work shows results across more diverse domains and studies the importance of supervised learning with multi-domain data and RL finetuning. Architecturally, \gea is different from OpenVLA~\cite{kim2024openvla} in that it uses a learned multi-embodiment action tokenizer as opposed to a uniform discretization, which prior work has demonstrated to perform better~\cite{szot2024grounding}. Moreover, works have explored the value of finetuning \mllms as UI agents~\cite{bai2024digirl,you2025ferret,gur2023real,furuta2023multimodal,nakano2021webgpt}. While \gea explores adapting \mllms as policies, there are also other ways to leverage \mllms such as via reward models~\cite{ma2024dreureka,ma2023eureka}, world models~\cite{wu2023unleashing}, or environment generation~\cite{yang2024holodeck}.

\looseness=-1 More broadly, prior work has demonstrated how LLMs can be used for agents that can reason and interact. Various works focus on training LLM agents through specific pipelines~\cite{zhang2024agentohana}. Similar to how \gea finetunes for decision-making capabilities not present in the base LLM, other works demonstrate the value of finetuning LLMs for reasoning and problem-solving capabilities~\cite{xie2024monte,singh2023beyond}. Prior works also demonstrate the value of finetuning LLMs with self-generated data~\cite{luo2024improve}, mistakes from the LLM~\cite{setlur2024rl}, and with RL~\cite{kumar2024training}. Likewise, \gea shows the importance of using such RL training, beyond just supervised learning, to create a capable agent.

%% file: sections/method.tex
\begin{figure*}[th!]
  \centering
  \includegraphics[width=0.9\textwidth]{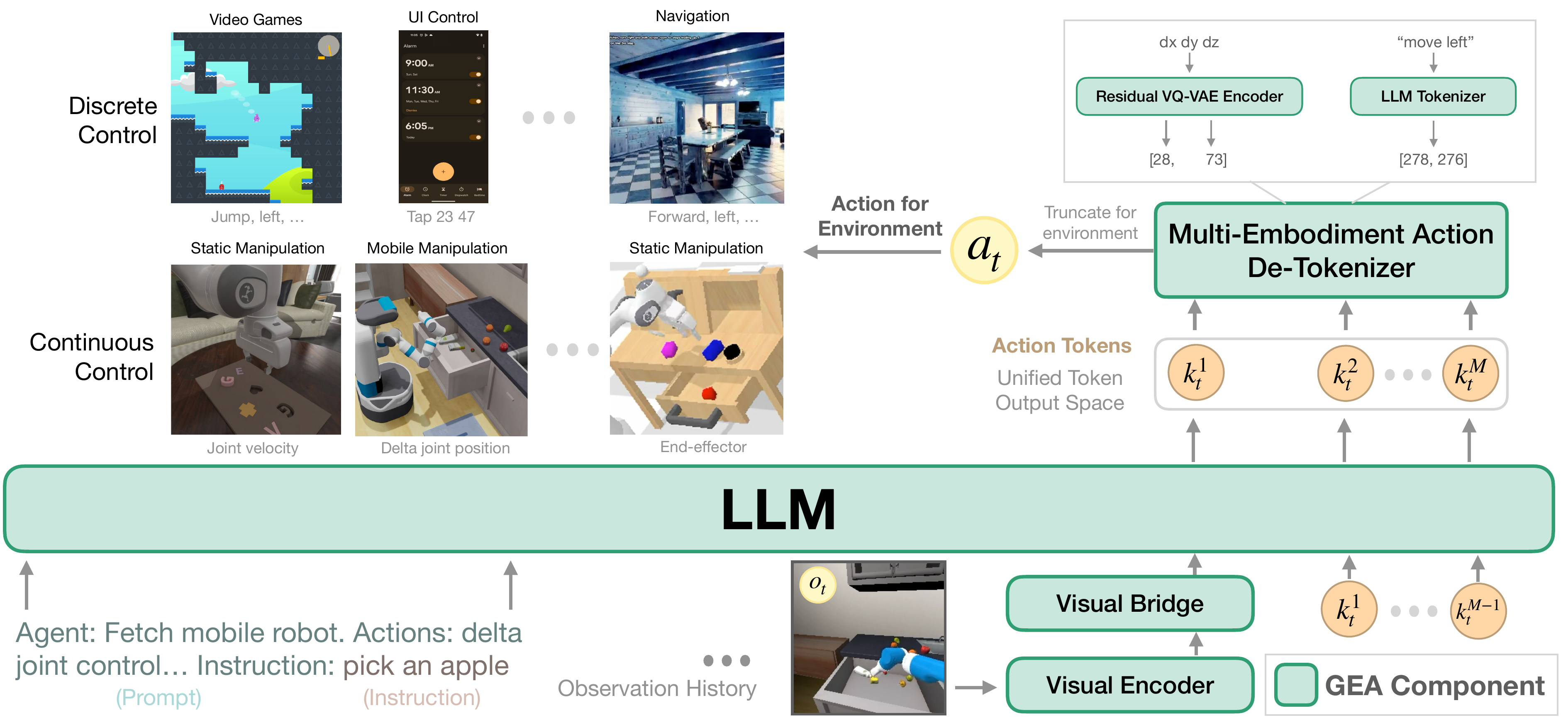}
  \caption{
    \looseness=-1 \gea utilizes a pretrained \mllm together with a multi-embodiment action tokenizer to enable a generalist agent to operate across a wide range of domains, embodiments, and action spaces. \gea takes as input information about the embodiment and desired task with the embodiment prompt and instruction and the observation visuals (bottom). It produces a sequence of action tokens in the LLM vocabulary, which are decoded by the multi-embodiment action detokenizer into an action for the appropriate embodiment and action space.
  }
  \label{fig:method}
\end{figure*}

\section{\geafull}
\subsection{Problem Settings}\label{sec:preliminaries}

We focus on language-specified tasks with visual observations. Specifically, we consider the goal-specified Partially-Observable Markov Decision Processes (POMDPs)~\cite{Bel} with observation space $ \mathcal{O}$, action space $ \mathcal{A}$, goal space $ \mathcal{G}$, and a reward model $R$. For brevity, we omit other elements of the MDP. 
In our settings, $ \mathcal{G}$ is represented by a textual description of the task to solve. 
Observations consist of RGB images from the agent, which can either be from an agent camera in the case of embodied AI applications or screenshots in the case of video games or UI interactions.

We consider a diverse set of environment types, which we refer to as \emph{domains} (see Table \ref{tab:datasets} for examples).
These domains specify a diverse set of action spaces spanning various robotic control spaces, high-level primitives, and computer UI interfaces. 
Our goal is to learn one policy that operates over a number of environments, which we denote by $\mathcal{M}_i=(\mathcal{O}_i, \mathcal{A}_i, \mathcal{G}_i, R_i)$ for each environment $ i \in  \mathcal{E}$.

\subsection{\gea Architecture}
\label{sec:arch} 
The \geafull (\gea) performs tasks by producing actions that are executed in an environment conditioned on observations, past actions, and a task description. More formally, for timestep $ t$ in environment $ \mathcal{M}_i$, the \gea model takes as input an environment specific prompt $ P_i$, followed by a task instruction $ I \in \mathcal{G}_i$ and up to $ c$ interleaved previous observations and actions $ o_{t-c}, a_{t-c}, \dots , o_t$ from $ \mathcal{O}_i$ and $ \mathcal{A}_i$. From these inputs, the \gea model predicts action $ a_t \in \mathcal{A}_i$ to execute in the environment.
The prompt provides information about the environment and embodiment to control. The task instruction is a natural language description of the task the agent is to execute.

\textbf{Multi-Embodiment Action Tokenizer.} We study \geafull (\gea) adapted from an \mllm. As \mllms naturally consume only text and images and generate only text, we follow the findings of \citet{szot2024grounding} to modify the LLM vocabulary to represent actions. First, we represent all actions across $\{\mathcal{M}_i\}$, $i\in\mathcal{E}$ with two vocabularies: $ V_\textrm{disc}$ for discrete actions and $V_\textrm{cont}$ for continuous actions, so that the final vocabulary is $V = V_\textrm{disc}\cup V_\textrm{cont}$. A \textit{discrete} action is described by language, and then this language is tokenized into a sequence of text tokens representing this action. $V_\textrm{disc}$ is defined as all such text token sequences representing the discrete actions.

As continuous actions are not readily expressed as text, we use a learned action tokenizer that maps each continuous action into a sequence of new tokens, whose vocabulary we denote by $V_\textrm{cont}$. Details of how we train this tokenizer are in \Cref{sec:tokenizer}. We replace the $|V_\textrm{cont}|$ most infrequently used tokens in the original LLM vocabulary with $V_\textrm{cont}$.

\section{Training}

\gea starts from a base \mllm and first trains a continuous action tokenizer. As depicted in \Cref{fig:training}, the \mllm is adapted to \geabase by supervised finetuning on embodied experiences. Next, \geabase is adapted to the full \gea model through supervised and reinforcement learning. 

\subsection{Base \mllm}
The primary consideration for selecting a base model beyond its inherent vision-language strength is its ability to scale to long contexts as embodied data consists of long trajectories of interleaved observations and actions. We thus build \gea off \lov~\cite{li2024llava}, a model specifically trained to handle sequences of images through interleaved image/text pairs and videos which extends to \gea operating over a history of observations.

\subsection{Continuous Multi-Embodiment Tokenizer}
\label{sec:tokenizer}

To obtain a vocabulary $V_\textrm{cont}$ for continuous actions and a tokenizer/de-tokenizer for these actions, we follow \citet{szot2024grounding} and train a Residual VQ-VAE (RVQ)~\cite{lee2022autoregressive} model over action vectors. The RVQ model is a variational autoencoder that leverages a sequence of discrete embeddings to represent the data. Specifically, it encodes an action $a$ as a sequence of $M$ tokens $k_1(a), \ldots, k_M(a)$, where each token denotes a code from a learned vocabulary $V_\textrm{cont}^m$, for $m \in {1, \ldots, M}$. A key feature of RVQ is that the $m^\textrm{th}$ vocabulary is trained to encode the residual of the action after it has been encoded with vocabularies ${1, \ldots, m-1}$. This hierarchical encoding makes RVQ effective in precisely representing continuous actions with a minimal number of discrete tokens.
The final continuous action vocabulary used by \gea is the union of the RVQ vocabularies $V_\textrm{cont}=\bigcup_m V_\textrm{cont}^m$.

The main distinction from \citet{szot2024grounding} is that we train a single tokenizer/de-tokenizer across all continuous action spaces. As shown in \Cref{tab:datasets}, these spaces cover a variety of robotic control types, including end-effector, joint velocity, and joint position control. To facilitate training a unified RVQ, we pad all action vectors to the maximum action dimension. During inference, we decode the predicted action tokens and then truncate the output to match the dimensionality of the specific embodiment's action space. We use 2 codebooks each with 512 tokens and a token vector dimension of 1024. Additional details on action tokenization are in \Cref{supp:multi-emb-tok}.

\begin{table*}[th!]
  \centering
  \resizebox{1.0\textwidth}{!}{
  \input{tables/datasets}
  }
  \caption{
    Overview of the embodied datasets used for training \gea. The actions in each dataset can be either discrete or continuous with a specific control space for the continuous actions. The embodiment type describes the agent being controlled. Each trajectory in a dataset refers to a sequence of images and actions. The data source refers to the collection method for these trajectories.
  }
  \label{tab:datasets}
\end{table*}

\subsection{Stage 1: Supervised-Instruction Finetuning}
\label{sec:method-sft} 

\begin{figure}[t] 
    \centering
    \includegraphics[width=1.0\columnwidth]{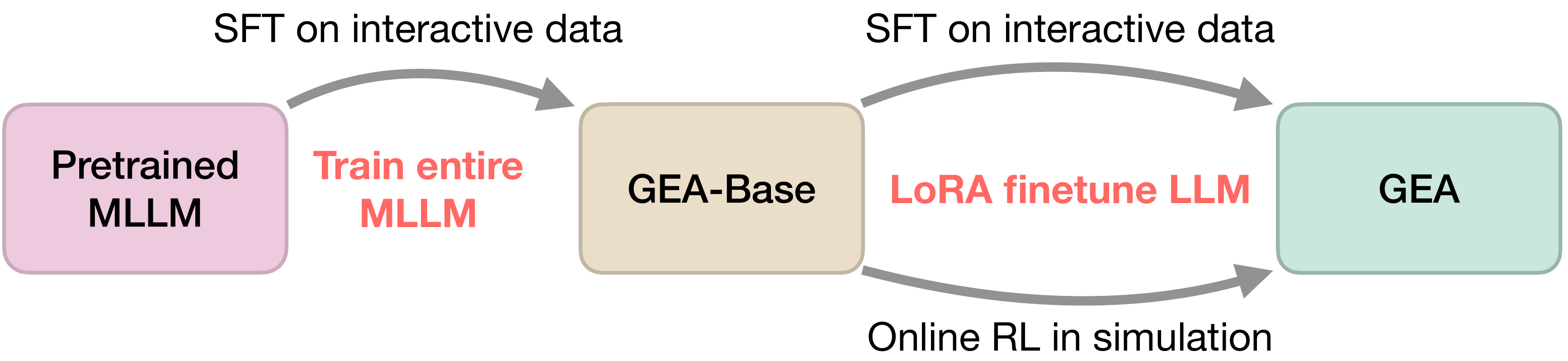}
    \caption{
      \gea training stages. First, a \mllm is adapted to \geabase by finetuning the entire \mllm with SFT on interactive data. Next, \geabase is finetuned jointly with online RL (PPO) and SFT on the original data with LoRA.
    }
    \label{fig:training}
\end{figure}

The first step of \gea is to use supervised-instruction finetuning (SFT) to adapt the base \mllm for embodied decision-making (left side of \cref{fig:training}). We use a collection $\mathcal{D} = \bigcup_{i\in\mathcal{E}} \mathcal{D}_i$ of demonstration datasets from all environments $\mathcal{E}$. 
During this stage, we use a standard cross-entropy loss over actions in the case of interactive data or responses in the case of vision-language data.
As is typically done in \mllm training, we maximize the negative log-likelihood of predicting these output tokens for each example:
\begin{equation}\label{eq:sft-loss}
\setlength{\belowdisplayskip}{3pt} \setlength{\belowdisplayshortskip}{3pt}
\setlength{\abovedisplayskip}{3pt} \setlength{\abovedisplayshortskip}{3pt}
  \mathcal{L}_{\textrm{SFT}}(\mathcal{D}) = - \!\!\!\!\!\!\!\!\!\!\!\!\sum_{(I, o_{t-c:t}, a_{t-c:t}) \in\mathcal{D}}\!\!\!\!\!\!\!\!\!\!\!\!\!\!\!\! \log p(a_t|P, I, o_{t-c}, a_{t-c}, \dots, o_{t})
\end{equation}
We then train the base \mllm over all the above datasets with SFT to obtain the \geabase model. 

\textbf{Training Details.} The entire \geabase model is initialized from the base \mllm. We train for 75k updates using AdamW~\cite{loshchilov2017decoupled} with cosine learning rate decay and linear learning rate warmup for the first $ 10\%$ of training steps; learning rate of $ 1e^{-5}$; global batch size of 256 and an observation context length of $ c =3$. Training takes around 2 days on 8 nodes of 8xH100 GPUs (see \Cref{supp:sft}).

\subsection{Stage 2: Online Reinforcement Learning}
\label{sec:rl-training}
While the previously described SFT training process produces a capable \geabase agent, it is only trained on a limited set of expert trajectories, which rarely demonstrate diverse behaviors like error recovery. 
Hence, we propose to utilize online RL for some of the environments. In a second stage of training, we continue to train \geabase with RL in addition to SFT to obtain the final \gea model (right side of \cref{fig:training}). 
For online RL, we train the \geabase agent with PPO~\cite{schulman2017proximal}, whose optimization loss is denoted by $\mathcal{L}_\textrm{PPO}(\mathcal{M}_i)$ for each environment in which we have a simulator $i\in \mathcal{E}_\textrm{PPO}\subset \mathcal{E}$. We combine this objective with the SFT objective from \cref{eq:sft-loss} to obtain the final \gea objective:
\begin{equation}\label{eq:final-loss}
    \mathcal{L}_{\text{GEA}} = \sum_{i\in \mathcal{E}_\textrm{PPO}}\mathcal{L}_\textrm{PPO}(\mathcal{M}_i) + \lambda\sum_{i\in \mathcal{E}}\mathcal{L}_\textrm{SFT}(\mathcal{D}_i)
\end{equation}
where we weight the SFT loss by $\lambda=0.1$ to emphasize the RL loss. This RL+SFT training stage on the \geabase model produces the final model we refer to as \gea.

\textbf{PPO Details.} The \gea value function for RL is an MLP network which is initialized from scratch. It takes as input the \mllm final layer activations at the observation token step just before the action tokens and an average pooled representation of the visual embeddings from the \mllm vision encoder. The critic also optionally takes any privileged state information about the task since the critic is only used during training, and not during inference. 

To stabilize RL training across numerous environments, we use PopArt~\cite{van2016learning} return normalization to account for the diverse reward distributions across these environments. Since the output space of the LLM can consist of many possible tokens, we use constrained decoding to force the autoregressive action sampling to be within the action space for the environment. For continuous action tasks, this amounts to constraining the output to the learned continuous action tokens. For the discrete control tasks this amounts to constraining the output to the valid language actions (for example, "pick apple" or "right"). To account for differences in the valid distribution of actions per environment, we normalize the entropy for PPO so a single entropy coefficient can apply to all tasks.

\textbf{Training Details.} Since \geabase already obtains some success, and RL introduces GPU memory overhead via environments simulated on the GPU, we use LoRA~\cite{hu2021lora} to finetune the LLM while freezing all other components. All environments use a rollout length of 128, a learning rate of $ 3e^{-4}$, an entropy coefficient of $ 1e^{-4}$, and a value function learning loss of $ 1.5e^{-4}$. RL finetuning uses 8 nodes of 8xH100 GPUs with 4 parallel environments per GPU. Environments are partitioned per node into 3 of the GPUs running HabPick, 3 running Procgen, and 2 running LangR. The SFT per-device batch size is 2. We train for 100M cumulative steps across all tasks, which takes around 1 day. Full RL training details are in \Cref{supp:rl}.

%% file: tables/datasets.tex
\begin{tabular}{cccccc}
\toprule
\textbf{Dataset Name}           &\textbf{Domain} & \textbf{Action Type}& \textbf{Embodiment Type} & \textbf{\# Trajs} & \textbf{Data Source} \\
  \toprule
  OpenX~\cite{o2023open} & Static Manip & Cont. Various & 22 Various Robots &  1.2M & Various \\
  Meta-World~\cite{metaworld} &Static Manip. & Cont. EE+Gripper&      Sawyer & 45k &  Scripted \\
  CALVIN~\cite{mees2022calvin} &Static Manip. & Cont. EE+Gripper&      Franka Arm &  18k&  Human \\
  Maniskill~\cite{gu2023maniskill2} &Static Manip. & Cont. Joint Velocity &   ROKAE xMate3Pro    &  5k &  Motion Planner \\
  Habitat Pick~\cite{szot2021habitat} &Mobile Manip. &        Cont. Joint Position + Base       & Fetch & 50k& RL Expert \\
  Habitat Place~\cite{szot2021habitat} &Mobile Manip. &        Cont. Joint Position+Base        & Fetch & 50k& RL Expert \\
  Habitat Nav~\cite{szot2021habitat} & Navigation & Cont. Velocity &   Fetch    &  13k &  Shortest Path \\
  BabyAI \cite{babyai_iclr19}& Navigation & Discrete &   Virtual    &  50k &  Shortest Path \\
  LangR~\cite{szot2023large} & Planning & Discrete & Fetch &  150k & RL Expert \\
  Procgen~\cite{cobbe2020leveraging} & Video Games & Discrete & Virtual &  320k & RL Expert \\
  Atari~\cite{bellemare2013arcade} & Video Games & Discrete & Virtual &  286k & RL Expert \\
  AndroidControl \cite{li2024effects} & UI Control & Mixed & Virtual &  14k & Human \\
  \midrule
\end{tabular}

%% file: sections/environments.tex
\begin{table*}[th!] 
  \centering
  \input{tables/main}
  \caption{
  Zero-shot generalization of \gea to new tasks in terms of success rate \% and in the video games tasks \% of expert performance. We compare against prior works consisting of domain specialists (with superscript ``S") that are trained on only data from that benchmark and domain generalists (with superscript ``G") that are trained on data from several benchmarks. \textbf{Bold indicates best}, \underline{underline close second}, and \textcolor{NoComp}{gray coloring} that the method assumes access to additional input modalities like pointcloud or ground truth simulator state, meaning it is not a fair comparison to \gea. The ``Prior Work" column also gives details about how the methods were trained (IL or RL) and if they assume additional input modalities. The ``\# Tasks" column gives a general indication of the number of distinct evaluation settings with a ``*" indicating each task also has diverse language instructions. 
  }
  \label{table:task-gen} 
\end{table*}

\section{Datasets and Environments}
\label{sec:environments}
We use a diverse set of domains with associated environments and datasets (see \Cref{tab:datasets}). This section introduces these domains followed by an explanation of how we use them in stage 1 and 2 of our training procedure.

\textbf{Static Manipulation.} These datasets are of a fixed robot manipulator interacting with objects. Some of these datasets are simulated table top interactions with rigid-body objects such as Meta-World~\cite{yu2019meta}, CALVIN~\cite{mees2022calvin} and Maniskill~\cite{gu2023maniskill2}. We also leverage a large dataset of interactions on real robot platforms~\cite{o2023open}. These datasets span a variety of control spaces in end-effector and joint control. The camera is typically mounted in a static position so that the workspace and robot arm are fully visible.

\textbf{Mobile Manipulation.} We also investigate setups where the robot manipulator moves via a mobile base. We use the object rearrangement tasks from the Habitat platform~\cite{habitat19iccv} for datasets in these tasks. These datasets cover object picking and placing tasks where the robot starts up to 2 meters away from the object. The robot has to coordinate moving its base and arm to successfully pick up the object. These datasets involve first person egocentric cameras.

\textbf{Navigation.} We also use datasets of navigation in isolation. We use datasets of simulated robot navigation in Habitat. We also use navigation in grid-world environments from BabyAI~\cite{babyai_iclr19}. Both datasets were collected with shortest path experts.

\textbf{Video games.} We use datasets from two standard benchmarks for decision making in video games,  Procgen~\cite{cobbe2020leveraging} and Atari~\cite{bellemare2013arcade}. Both datasets were collected by RL agents which were separately trained to solve each individual game. 
We convert these tasks to be language-conditioned by providing the game name along with a short description of the game's objective and rules. We only train on successful trajectories.

\textbf{Planning.} We use a dataset of successful episodes in the LangR dataset~\cite{szot2023large}. In this task the agent has to select between skill primitives that accomplish long horizon language-specified rearrangement tasks for a home robot.

\textbf{UI Control.} We use the AndroidControl~\cite{li2024effects} dataset of UI interaction in Android devices spanning 833 apps. The actions are combinations of tap actions specified by screen coordinates and text typing.

\textbf{Vision language instruction data.} To improve generalization of the model, we also include data used for training the original \mllm, which prior work has found is useful when finetuning \mllms as control policies~\cite{brohan2023rt}. We used the following datasets of text and images without any actions: VQAv2~\cite{goyal2017making}, OKVQA~\cite{marino2019ok}, A-OKVQA~\cite{schwenk2022okvqa}, GQA~\cite{hudson2019gqaan} and the LLaVA-Instruct-150k dataset~\cite{liu2024improved}.

\textbf{Stage 1 Training: SFT Data.} To obtain embodied data for Stage 1 Training (see Sec.~\ref{sec:method-sft}), we collect a large dataset of language-conditioned behaviors from all of the above domains consisting of 2.2M trajectories. All trajectories are successful examples of language-conditioned behaviors with visual observations. The data is from diverse collection sources such as human demonstrations, RL-based policies, and motion planners. The dataset is diverse and spans thousands of distinct tasks and many embodiments (see \Cref{supp:dataset} for full details).

\textbf{Stage 2 Training: RL Environments.} For Stage 2 online RL (see \cref{sec:rl-training}) we use environments from the three domains of Habitat Pick~\cite{szot2021habitat}, Language Rearrengement (LangR)~\cite{szot2023large}, and Procgen~\cite{cobbe2020leveraging}. Thus, we define $\mathcal{E}_\textrm{PPO}=\{\textrm{HabPick}, \textrm{LangR}, \textrm{ProcGen}\}$. Habitat Pick and LangR are simulated in the Habitat platform~\cite{habitat19iccv} and have reward functions for achieving and making progress towards the goal. In Procgen, we use all 16 games for RL and use the game specific reward functions (see \Cref{supp:rl} for full details).

%% file: tables/main.tex
\begin{tabular}{P{0.20\textwidth}P{0.10\textwidth}L{0.20\textwidth}P{0.10\textwidth}P{0.20\textwidth}}
\toprule
 & \textbf{\gea} & \textbf{Prior Work} & \textbf{\# Tasks} & \textbf{Generalization Type} \\
\midrule
\textbf{Manipulation} &  &  &  &  \\
Meta-World & \textbf{  94.7  } & 84 \textcolor{NoteColor}{MLLM+IL} \cite{szot2024grounding}\textsuperscript{S} \newline 87.0 \cite{reed2022generalist}\textsuperscript{G}  & 45 & object positions \\
\rowcolor{Gray} CALVIN (ABC \textrightarrow D) & \textbf{  90.0  } & 82.4 \textcolor{NoteColor}{MLLM+IL} \cite{li2023vision}\textsuperscript{S} \newline \textcolor{NoComp}{92.2} \textcolor{NoteColor}{IL+pointcloud}\cite{ke20243d} & 34* & instructions, background \\
Maniskill &  13.6  & 6.5 \textcolor{NoteColor}{IL} \cite{gu2023maniskill2}\textsuperscript{S} \newline \textbf{47.8} \textcolor{NoteColor}{IL+PPO} \cite{gu2023maniskill2}\textsuperscript{S} & 5 & object positions \\
\rowcolor{Gray} Habitat Pick & \textbf{  82.5  } & 29 \textcolor{NoteColor}{IL} \cite{szot2024grounding}\textsuperscript{S} \newline \textcolor{NoComp}{81.0} \textcolor{NoteColor}{RL + sim state}\textsuperscript{S} & 20 & house \\
Habitat Place & \textbf{  93.5  } & \textcolor{NoComp}{95.5} \textcolor{NoteColor}{RL + sim state}\textsuperscript{S} & 10 & house \\
\hline
\textbf{Video Games} &  &  &  &  \\
Procgen & \textbf{  44.0  } & 25 \cite{mediratta2023generalization}\textsuperscript{S} & 16 & background \\
\rowcolor{Gray} Atari &  32.7  & 31 \cite{reed2022generalist}\textsuperscript{G} \newline \textbf{85} \textcolor{NoteColor}{Offline RL} \cite{lee2022multi}\textsuperscript{S} & 44 & none \\
\hline
\textbf{Navigation} &  &  &  &  \\
Habitat Nav & 62.5  & \textbf{72} \cite{szot2021habitat}\textsuperscript{S} & 10 & house \\
\rowcolor{Gray} BabyAI & \underline{91.1} & \textbf{93.2} \cite{reed2022generalist}\textsuperscript{G} & 17* & instructions, grid state \\
\hline
\textbf{UI Control} &  &  &  &  \\
AndroidControl & \textbf{  57.3  } & 45 \textcolor{NoteColor}{GPT-4o+SoM} \cite{yang2023set}\textsuperscript{G} & 35* & instructions \\
\hline
\textbf{Planning} &  &  &  &  \\
LangR & \underline{50.0} & \textbf{51} \textcolor{NoteColor}{MLLM+RL}\cite{szot2024grounding}\textsuperscript{S} & 10* & instructions, house \\
\hline

\end{tabular}

%% file: sections/experiments.tex
\section{Empirical Evaluation}
\label{sec:experiments} 
We empirically demonstrate the ability of \gea as a generalist agent that can generalize to new instructions and settings across diverse embodiments and domains. We assess the role of the RL training in achieving this goal. In ablations, we study the impact of scaling data between multiple domains, compare RL to other forms of policy collected data, the impact of the the base MLLM. 

\subsection{\gea Generalization Capabilities}
\label{sec:gea-gen} 

In this section, we evaluate the generalization capabilities of the final \gea model. We use the associated benchmarks from the datasets in \Cref{tab:datasets}, which span manipulation, navigation, video games, UI control, and planning. These benchmarks evaluate agents in new settings not present in the training data, such as new object positions, scenes, visual backgrounds, tasks, or instructions. All benchmarks specify evaluation instructions in natural language and require the agent to operate from visual observations. 

We report the ``online" performance of \gea, meaning we evaluate its performance in an interactive simulator. The only exception is AndroidControl, where we instead check the correspondence with a ground truth test trajectory~\cite{rawles2024androidinthewild}. Each benchmark also evaluates agents over many distinct tasks. For example, in the Procgen video game benchmark, we report the average Procgen performance over all 16 Procgen games each of which is an entirely different video game. Full evaluation details are in \Cref{sec:eval-details}. 

\begin{table}[t]
  \centering
  \resizebox{1.0\columnwidth}{!}{
    \input{tables/all_online}
  }
  \caption{
    Effect of stage 2 RL training on \geabase (7b model). Tasks included in RL training increase their generalization performance. Performance on the remaining tasks slightly increases thanks to continued SFT training.
  }
  \label{table:all-online} 
\end{table}

We seek to comprehensively frame the empirical performance of \gea relative to prior work on our evaluated benchmarks. First, in all benchmarks, we use only images as observations without using any privileged information such as the simulation state or additional observations such as 3D point clouds. Second, we evaluate our single \gea model across all environments, which is referred to as a generalist. Some of the approaches we compare against are trained on data from a single environment, and we refer to those as specialists. In other comparative approaches, the split between train or test data is unclear. For example, Gato~\cite{reed2022generalist} is a generalist model like \gea, yet evaluates and trains on some less diverse tasks with their own training datasets, which are not released. \Cref{sec:baselines} discusses these connections in detail.

 \Cref{table:task-gen} summarizes the comparative evaluation. \gea excels at manipulation tasks, either exceeding or matching the performance of specialist models. For example, in Meta-World \gea greatly outperforms both specialist and generalist models trained for this task with a $ 7\% $ absolute increase relative to the best-performing baseline.
In CALVIN, \gea outperforms a variety of recent specialist models. \gea also performs closely to the specialist 3D Diffuser Actor method~\cite{ke20243d}, which uses a manipulation-specific action representation of end-effector key points and uses a depth camera to represent the scene as a 3D feature cloud. \gea only uses the third PoV RGB camera and does not use an action or observation space specific to table-top manipulation. 
\gea outperforms the baseline in Habitat Pick and closely matches it in Habitat Place despite these baselines being trained with the ground truth simulator state. In Maniskill, the difficult, often occluded, camera view results in low overall success rates, yet \gea outperforms other results that only use IL. However, \gea is outperformed by methods that use RL in this benchmark. 

In video game benchmarks, \gea outperforms the specialist model baseline in Procgen. In Atari, \gea outperforms the generalist Gato model~\cite{reed2022generalist}. However, in Atari, \gea is outperformed by Multi-Game DT~\cite{lee2022multi}, which uses offline RL from suboptimal demonstrations. In Atari, \gea does not learn with offline or online RL.

In the BabyAI navigation benchmark, \gea is close in performance to Gato~\cite{reed2022generalist} despite \gea using RGB renderings of the top-down view rather than any underlying state information and 100x fewer demonstrations for this environment. In Habitat Nav, \gea underperforms an RL-trained expert. This gap could be due to the context of \gea only consisting of the previous three observations, which could limit its ability in partially observable settings. In UI control, another discrete action task, \gea outperforms GPT-4o~\cite{OpenAI} with set of marks~\cite{yang2023set} generated by a UI detection model. This demonstrates that \gea benefits from being specialized for interactive decision-making even against a powerful LLM and specialized perception system. Finally, in the discrete control benchmark of the LangR planning task, \gea closely matches the performance of a specialist baseline, which trains on only this task with RL. 

\begin{table}[t]
  \centering
  \resizebox{1.0\columnwidth}{!}{
    \input{tables/llm_pretraining}
  }
  \caption{
  We present results using \lovsmall as the MLLM base (row 1) as well as training with only domain specific data (row 2). We also present results for a transformer of the same architecture but only the LLM subnet is initialized (row 3), or the visual encoder (row 4), or neither (row 5). 
  }
  \label{table:llm-pretraining} 
\end{table}

\textbf{Comparative Benefit of RL to SFT.} Training with RL was important to achieving these strong results. \Cref{table:all-online} compares the performance of \geabase, which is only trained with SFT on expert demonstrations, and \gea, which is additionally trained with a second stage of RL. The results show that the success rate of \gea greatly increases on Habitat Pick, Procgen, and LangR, which are the environments where RL was used. Furthermore, the average success rate across all the other tasks is not influenced by RL due to the continued joint SFT training.

\subsection{\gea Training and Model Analysis}
\label{sec:gea-analysis} 

In this section, we explore the relationship between the generalist capabilities of \gea and its training data. 
We analyze the role of embodied SFT data for adapting the \mllm for interaction and the importance of online data.
We also evaluate the importance of the base \mllm. 
For all results in this section, we train all models with 32\% of the original embodied SFT data and 40k updates to ease the computational burden of the analysis. We also evaluate on the tasks from \Cref{table:task-gen} with a reduced number of evaluation episodes with around 200 episodes per benchmark. \Cref{supp:analysis-dataset} describes this analysis evaluation and dataset in detail.

\textbf{Impact of Multi-Domain Data.} We evaluate training a generalist model on data from all diverse domains, versus training a model only on data from the particular target domain. Specifically, we train the smaller \lovsmall model on either data from a single domain (``Domain Specific") or data across all domains (``\geabase"). Comparing these options in \Cref{table:llm-pretraining} shows that across all the benchmarks, training with all the data is beneficial. However, the gain is smaller in some of the domains like Android Control and Procgen, likely due to less overlap with the other training domains as opposed to the wealth of manipulation data we train with.
\Cref{supp:transfer} contains a more granular investigation into the pairwise transferability between each dataset and also supports this conclusion that \gea benefits from multi-domain data.

\begin{figure}[t] 
  \centering
  \includegraphics[width=1.0\columnwidth]{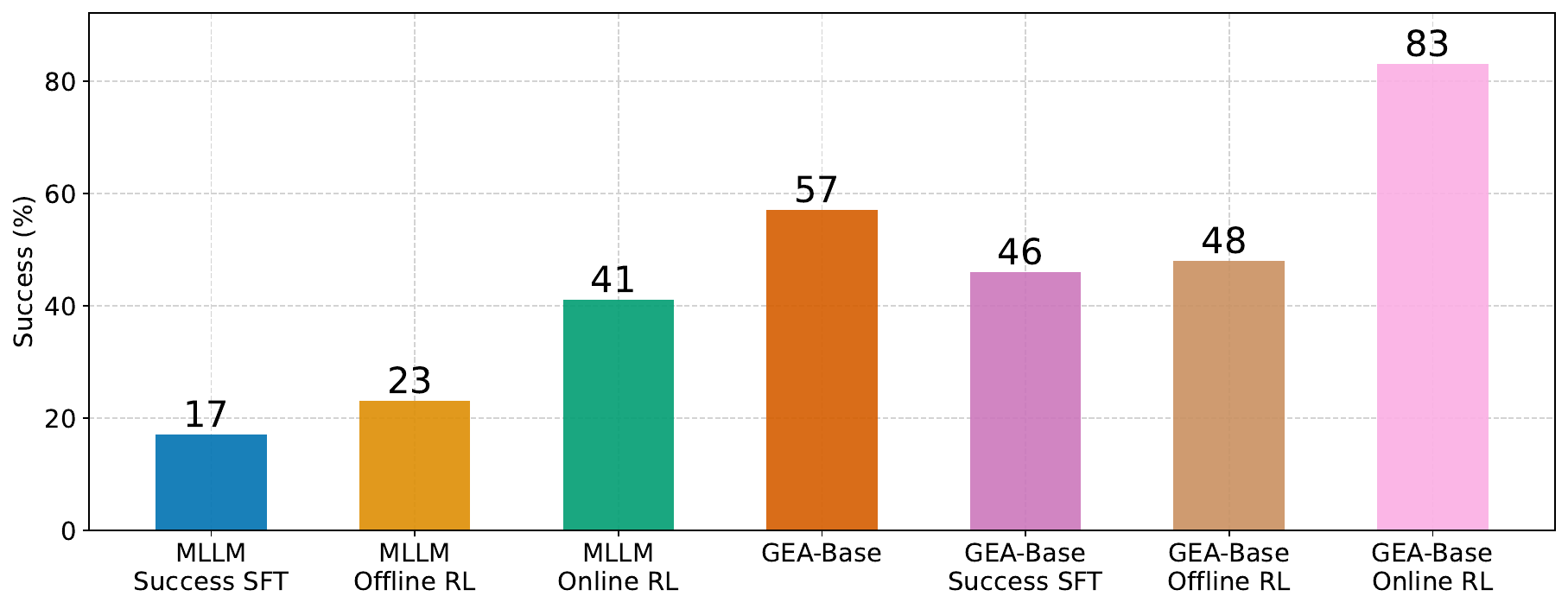}
  \caption{
    Online learning in Habitat Pick. MLLM methods finetune LLaVA-OV while other methods finetune \geabase. 
  }
  \label{fig:habpick-online}
\end{figure}

\textbf{Impact of Policy-Collected Data.} Next, we explore the role of learning from data sources beyond SFT on expert demonstrations. While \geabase is a capable embodied policy, it is only trained on successful demonstrations. These demonstrations rarely exhibit recovery behaviors or robustness to non-expert behaviors. 
Unlike typical \mllm applications such as visual question answering, in interactive tasks, agents trained with expert data can suffer from the problem of ``covariate shift" where small agent errors cause the observation distribution to shift from the expert's and for errors to compound ~\cite{ross2011reduction}. 

We analyze how \geabase can be trained with additional data to improve performance in the Habitat Pick task and compare the following alternatives. First, \textit{\geabase Success SFT} collects 10k successful examples in the environment with the \geabase policy and then trains on these successes with supervised learning. \textit{\geabase Offline RL} collects 10k trajectories consisting of both successes and failures, both labeled with the dense Habitat Pick reward, and then trains on these with the IQL offline-RL algorithm~\cite{kostrikov2021offline}. \emph{\gea Online RL} finetunes \geabase with PPO, which leverages online interactions with the simulator like the \gea Stage-2 training (but omits the joint SFT loss). We again use the smaller base \lovsmall model for these experiments.
\Cref{fig:habpick-online} shows the results of these variations.

A main takeaway message is the \textbf{strong impact of online RL} on top of a finetuned \mllm, which increases the success of \geabase from $ 57\% $ to $ 83\%$, despite the latter being trained on 50k successful Habitat Pick demonstrations. Online RL outperforms both Success SFT and IQL offline RL, highlighting the need for online interactions.
It is worth noting that applying success SFT and offline RL on top of \geabase decreases the model's performance, which could be due to the lack of diverse data.

These results further show that \textbf{online RL is beneficial when applied on top of a finetuned model with domain data}. Online RL alone is unable to bring the performance of the base \mllm to \geabase without the stage 1 SFT. \Cref{supp:habpick-rl} explores this further and shows that \geabase is also far more sample efficient with RL than the base \mllm. This analysis demonstrates it is important for the \gea model to use RL in combination with SFT.

\begin{figure}[t] 
  \centering
  \includegraphics[width=1.0\columnwidth]{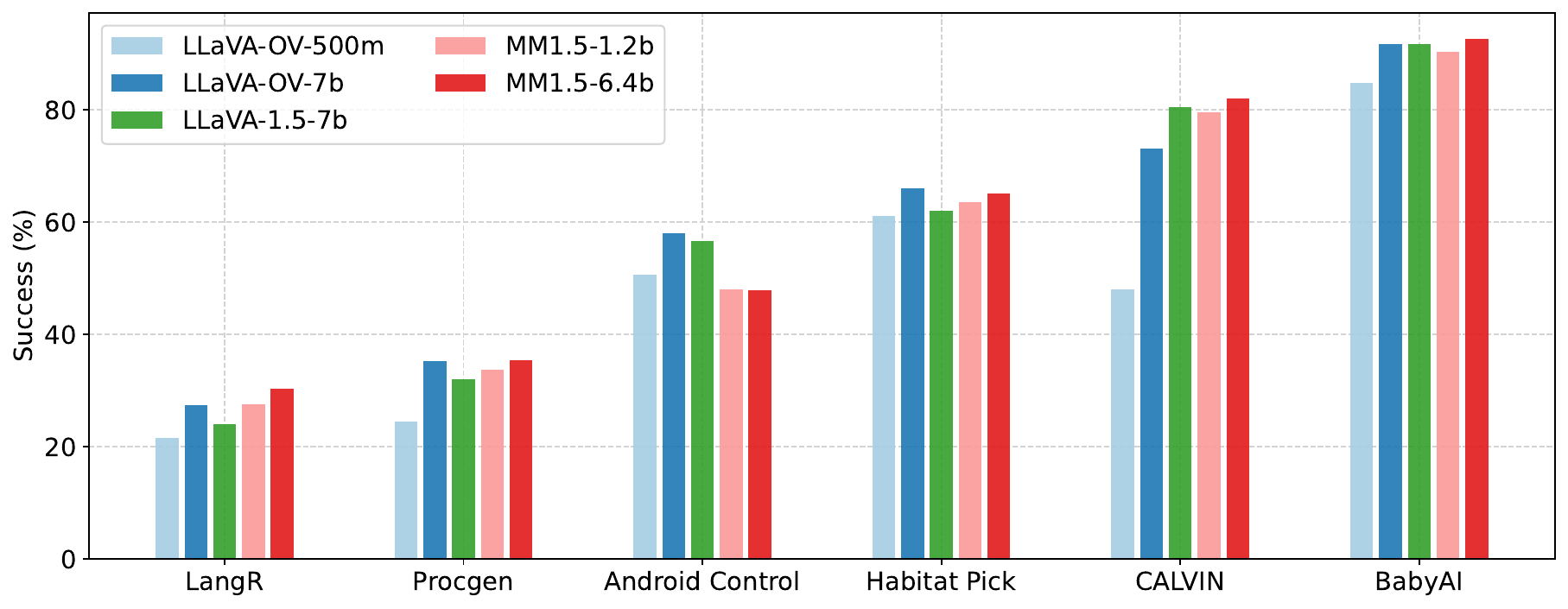}
  \caption{
    Analyzing the impact of training \gea with different base \mllms with different parameter counts. 
  }
  \label{table:llm} 
\end{figure}

\textbf{Impact of pretrained \mllm.} We assess the importance of the pretrained \mllm in the \gea architecture. To do so, we present results using a base model that has an architecture identical to the pretrained \mllm. However, instead of initializing the full model with \lov we initialize only the LLM or the visual encoder with the corresponding subnet weights.

The results in \Cref{table:llm-pretraining} show that the full \mllm has a substantial impact on performance. Although this isn't surprising, it is important to note that the visual encoder initialization seems to have a stronger impact on the final performance compared to the LLM. We conjecture that this is because the benchmarks require visual generalization, and training the \lov SigLIP visual encoder from scratch with only the embodied data is challenging.

Further, we analyze the impact of the base \mllm size on the \geabase. We use two classes of backbone models, \lov~\cite{li2024llava} and MM1.5~\cite{mckinzie2024mm1}, and use two sizes for each of them ($0.5B$ and $7B$ for the former, and $1.2B$ and $6.4B$ for the latter). As shown in \Cref{table:llm}, in our agentic applications \textbf{increasing model capacity leads to stronger performance, both within and also across backbone model class}. Interestingly, the class of models has little effect as we see very similar performance across multiple backbone models. The three different $7B$ models have been trained on different web-scale data. Nevertheless, their difference has a negligible impact on \gea.

Additionally, in \Cref{supp:model-var}, we show that \gea training and evaluation is robust to the random seed.

%% file: tables/all_online.tex
\begin{tabular}{P{0.20\columnwidth}P{0.20\columnwidth}P{0.20\columnwidth}P{0.20\columnwidth}P{0.20\columnwidth}}
\toprule
 & \textbf{Habitat Pick} & \textbf{Procgen} & \textbf{LangR} & \textbf{All Other} \\
\midrule
\gea & \textbf{  82.5  } & \textbf{  44.0  } & \textbf{  50.0  } & \textbf{  70.5  } \\
\geabase &  60.5  &  36.1  &  15.5  &  69.5  \\
\bottomrule

\end{tabular}

%% file: tables/llm_pretraining.tex
\begin{tabular}{rP{0.12\columnwidth}P{0.12\columnwidth}P{0.12\columnwidth}P{0.20\columnwidth}P{0.12\columnwidth}P{0.12\columnwidth}}
\toprule
 & \textbf{Habitat Pick} & \textbf{CALVIN} & \textbf{Procgen} & \textbf{Android Control} & \textbf{BabyAI} \\
\midrule
  GEA-Base &  \textbf{57.0}  & \textbf{  48.0  } &  \textbf{24.5}  & \textbf{  50.5  } & \textbf{  84.7  } \\
\hline
Domain Specific &  54.5  &  35.5  &  23.7  &  48.9  &  82.1  \\
\hline
Only LLM &  9.5  &  0.0  &  7.6  &  26.4  &  49.4  \\
\rowcolor{Gray} Only VisEncoder &  34.5  &  13.0  &  \textbf{24.5}  &  28.3  &  70.6  \\
None &  9.0  &  0.0  &  7.4  &  14.1  &  44.4  \\

\end{tabular}

%% file: sections/conclusion.tex
\section{Conclusion}

In this work, we studied how finetuning pretrained \mllms with large-scale embodied experience via expert trajectories and online RL unlocks their ability to act as \geafull. To interface with diverse embodiments, \gea uses a learned action tokenizer. We illustrate the importance of RL finetuning for \gea, which results in competitive results across a variety of domains spanning manipulation, video games, navigation, UI control, and planning. 

While \gea demonstrates impressive abilities across a wide diversity of tasks, it is still not at the level of a foundation model for decision-making like similar models for language and vision~\cite{achiam2023gpt}. \gea cannot control arbitrary embodiments and operate in arbitrary environments zero-shot. Furthermore, the performance of \gea in some domains, such as Maniskill, Atari, and AndroidControl, is far from perfect. Extending RL to these environments could be a solution. Future work can continue to scale \gea to more tasks to extend its generalist capabilities.

%% file: supp/method.tex
\section{Continuous Multi-Embodiment Tokenizer Details}
\label{supp:multi-emb-tok} 

As stated in \Cref{sec:tokenizer}, we train a Residual VQ-VAE (RVQ) model to convert continuous actions from diverse robot embodiments into a shared discrete token space. In our experiments, we use 2 codebooks, each with 512 tokens, and the token vector dimension is 1024. Actions are encoded into these latent codebooks via a 4-layer MLP with 4096 hidden dimensions per layer. The decoder uses the same MLP architecture but now inputs the 1024-dimension latent and outputs the padded action. Refer to \citet{lee2022autoregressive} for further details about the RVQ method. We map these 512 tokens for both of the 2 codebooks to tokens $ 30000 - 30512$ of the LLM vocabulary. Since these tokens are non-English tokens for all the LLMs we consider and all our tasks use English instructions, we use this same token range for all \mllm experiments. All actions are padded to be 21 dimensions during tokenization. During detokenization, the 21 dimension continuous vector is truncated from the right to fit the expected action dimension for the environment.

The RVQ tokenizer is trained with MSE loss using all datasets with continuous actions from the overall set of datasets used to train \gea described in \Cref{supp:dataset}. The commitment loss from the RVQ is weighted by $ 1.0$ relative to the MSE loss. The tokenizer is trained with a per-GPU batch size of 256 across 8 H100 GPUs. The model is then trained for 15k updates using the AdamW optimizer~\cite{loshchilov2017decoupled} with a cosine learning rate decaying to 0 at the end of training and a linear learning rate warmup schedule for the first $ 10\%$ of training. At this number of updates, the validation MSE loss on unseen actions had largely plateaued at $ \approx 0.0038$ averaged across all datasets and $ 0.002 - 0.007$ depending on the dataset. This trained RVQ tokenizer was then used to train and evaluate all models with continuous actions.

\section{SFT Training Additional Details}
\label{supp:sft} 

To limit the number of tokens per frame to be processed by the LLM, we use the ``video" encoding strategy from \lov. This does not use the AnyRes technique and also applies bilinear interpolation to reduce the number of tokens per image. This results in 196 tokens for each image after being processed by the visual encoder and downsampled. For the training sequence format described in \Cref{sec:arch}, we use the following prompt format where per-domain strings are manually defined and substituted into each bracketed statement.
\begin{lstlisting}[frame=single]
User:
Agent: <agent description>.
Actions:
Simulator: <simulation platform>
Camera: <camera details>
Instruction: <task instruction>

Agent: 
\end{lstlisting}
When forming training batches, we randomly sample trajectories, and then randomly sample a span of the appropriate context length from that trajectory, and then use the sampled span of observation and actions as an element of the data batch. In the case of the interactive data, the model is only trained to predict the action tokens; the loss for the prompt, instruction, and image tokens is masked out.

For the details about the datasets used in this training process, see \Cref{supp:dataset}.

\section{RL Training Additional Details}
\label{supp:rl} 

In this section, we list additional RL training details omitted from \Cref{sec:rl-training}. The value function is a 4-layer MLP with a hidden dimension of 2048. The value function takes as input a single vector from the mean-pooled visual tokens from the visual encoder, the final activation of the LLM for the observation, and any task-specific state information that is available. For LoRA finetuning, we use a value of $ r = 128, \alpha = 32 $, and a dropout value of $ 0.1$. We use GAE return estimation with $ \tau = 0.95$ and a discount factor of $ \gamma = 0.999$. 

We run RL with the Habitat Pick, LangR, and Procgen environments. Each GPU runs a different benchmark instance. Both Habitat Pick and Procgen are allocated to run on $ 50\%$ more GPUs than LangR because we found that LangR learns much faster with RL compared to the other tasks. Per benchmark variation, such as different episodes in Habitat Pick and different games in Procgen, are equally divided between the GPUs assigned to that benchmark. On each GPU, a parallelized set of 6 environments are running for batched environment experience collection. We update with 2 PPO epochs, 6 minibatches per epoch, and a clip parameter of $ 0.2$. We also use the AdamW optimizer for RL training but without any learning rate schedule. We now detail the RL considerations specific to each environment.

In Habitat Pick, we use the same environment details as from prior works using the same task \cite{szot2024grounding,harish2025reinforcement,habitatrearrangechallenge2022}. The task requires the agent to pick up a target object specified by name. Specifically, the agent starts within 2 meters of the target object and faces towards the receptacle the target object is on but with random noise $ \mathcal{N}(0, 1.57)$ applied to the direction of facing directly at the receptacle. The task ends in failure if the agent excessively collides with the scene, drops the object, or picks up the wrong object. The maximum number of steps per episode is 300 steps. We use the same reward as the RL-trained pick skill from \citet{szot2021habitat} where a dense reward is provided for moving the end-effector closer to the object, a positive bonus for picking up the object, and then negative penalties for collisions. We use the 50k training episodes from \citet{habitatrearrangechallenge2022}. Note that these training episodes used for RL training are distinct from the episodes used for testing, which are in unseen house layouts. 

For the LangR environment, we use the RL environment details from \citet{szot2023large}. The reward function for this environment involves a sparse reward for completing the task, subgoal rewards for completing individual parts of the task, and a slack penalty to encourage completing the task faster. This environment has a maximum of 32 steps per task. In LangR, memory is important because the agent must explore to find certain objects. We therefore increased the number of visual observations in the context to 16 for the LangR RL training. We achieve this by increasing the visual encoder bilinear interpolation factor to produce only 32 visual tokens per image observation. Implementing such environment-specific policy tweaks is simpler in the RL training phase because each GPU worker runs a distinct environment. 

For Procgen, we use the RL environment details from \citet{cobbe2020leveraging}. Importantly, we train over all 16 of the Procgen games at once using RL. We use the standard per-game reward functions.

%% file: supp/datasets.tex
\section{Dataset Details}
\label{supp:dataset} 

\textbf{MetaWorld}: 
We use the Metaworld simulator~\citep{yu2019meta} with the pre-defined MT-45 split, which consists of 45 training tabletop manipulation tasks using a Sawyer XYZ arm. Each of the 45 tasks has a language instruction associated to it, e.g. "Open the drawer", which we use in all experiments. As inputs to the generalist agent, we use RGB observations from the $\texttt{corner3}$ camera resized to~\gea resolution. We use no proprioceptive information.

The dataset consists of 500 trajectories from each of the 45 train tasks. To construct the train, validation and test datasets, we rollout the scripted expert policy 10 times for each task's starting state until the first success, to obtain sufficient data diversity. We then assign some of the starting states from the 45 training tasks to the validation set and the remaining trajectories to the training dataset.

\textbf{Atari}: We use the DQN replay dataset~\citep{agarwal2020optimistic}, which consists of 50 million transitions collected while training the DQN~\citep{mnih2013playing} algorithm on each of the 44 environments separately. Based on the findings of Multi-Game Decision Transformer~\citep{lee2022multi}, we construct the SFT training dataset to be the top-$10\%$ of the DQN replay data by trajectory returns, over 5 random seeds and 50 splits. We do not use $100\%$ of the data since it is suboptimal to include low-return trajectories in the SFT dataset. However, if one would like to run offline RL on the data, they should include all of the available data.

\textbf{BabyAI}: This dataset consists of 50k trajectories split equally between each of the BabyAI tasks from \citet{babyai_iclr19}. These trajectories are collected by the shortest path expert provided in the code release for \citet{babyai_iclr19}. The observations are top-down RGB renderings of the image at $ 336 \times 336$ resolution.

\textbf{Procgen}: We use the training datasets across all Procgen games provided from \citet{mediratta2023generalization}. This dataset consists of 10M observation-action transitions between all the games collected by a PPO expert policy that was individually trained on each game. These transitions amount to around 320k trajectories. 

\textbf{CALVIN}: We use the standard CALVIN $ ABC \rightarrow D$ dataset provided by \citet{mees2022calvin}, which are language-labeled task instructions collected by human teleoperation of the robot. This dataset consists of around 18k demonstrations. The observations are $ 200 \times 200$ RGB renderings from the robot head camera. Note that unlike prior work~\cite{li2023vision}, we do not use the gripper camera or any robot proprioception from this dataset. The actions in the dataset are 6D end-effector control for the relative orientation and position and the gripper state.

\textbf{LangR}: For this work, we collect 150k demonstrations for each of the 150k unique training episodes defined by \citet{szot2023large}. We collect this data by utilizing the RL-trained policy from \cite{szot2023large}. This policy achieves high performance on the training set of instructions (around $ 98\%$ success rate), and we collect 1 successful demonstration per training episode. The observations are $ 336 \times 336$ RGB images from the robot head camera. The actions are between 70 high-level skills that include picking up objects by name, navigating to receptacles, placing on receptacles by name, and opening and closing receptacles by name.

\textbf{Habitat Pick/Place}: We collect 50k demonstrations for each of these tasks via an expert policy trained with RL. We use the same setup as from the skill training in \citet{szot2021habitat} but operate from an object class to pickup rather than the original geometric goal task specification. This expert policy was trained with the ground truth simulator state, consisting of the relative position of the target object to the robot's end-effector. The expert is trained for 100M steps until convergence. The performance of both experts on the unseen episodes are displayed in \Cref{table:task-gen} as baselines. 

For the online learning experiments from \Cref{sec:gea-analysis}, we construct the training dataset as follows. We use the GEA-Base checkpoint to collect 10,000 trajectories with action sampling to allow for suboptimal trajectories to be added to the data. The rollouts policy has a success rate of about $50\%$, meaning that half of the trajectories can be used for SFT from successful trajectories, and all of them can be used for offline RL.
When training the SFT policy, we restrict the loss to be optimized only over successful trajectories, as SFT cannot learn from suboptimal data. In addition to observations and actions, we also log the reward values which are required for offline RL.

\textbf{Habitat Nav}: We collect 13k shortest path demonstrations of an agent navigating to receptacles by name in the house. The navigation is performed by an oracle shortest path agent. 

\textbf{Maniskill}: These datasets are from the released imitation learning datasets from \citet{gu2023maniskill2}. They are generated via a motion planning algorithm. We only utilize the RGB 3rd-person RGB camera. We generate data for the ``StackCube", ``PegInsertionSide", ``PlugCharger", ``PushCube", and ``PickCube" tasks. The dataset from \citet{gu2023maniskill2} has 1k demonstrations for each of these tasks.

\textbf{Android Control}: This is the dataset from \citet{li2024effects} consisting of 14k human demonstrations of using apps to accomplish UI control tasks. Each trajectory has a unique language instruction. The data spans 833 apps. The original images are $ 1080 \times 1920 $, corresponding to the phone screen size. Since we do not use any AnyRes techniques in the \mllm visual encoder, we resize the images to be square before inputting them into the \mllm visual encoder. Using AnyRes with image crops to account for these image aspect ratios could improve the performance of \gea. Actions are generally represented as the name of the action type (like ``tap", ``scroll", or ``input text") followed by the argument for that action where applicable. For tap action arguments, we discretize the original $ 1080 \times 1920$ screen into $ 50 \times 50$ patches. A tap action argument is represented as two integers representing the horizontal and vertical patch coordinates where the tap occurred on the screen. All these actions are represented as textual tokens and are separate from the continuous action tokens.

\textbf{OpenX}: The Open X-Embodiment dataset consists of numerous individual datasets spanning different robots. We include the following individual datasets from OpenX: Austin Buds dataset, Austin Sailor dataset, Austin Sirius dataset, Berkeley Cable Routing, CMU Stretch, DLR EDAN Shared Control, Fractal, IAMLab CMU Pickup Insert, Jaco Play, Kuka, UCSD Kitchen Dataset, UTAustin Mutex, Dobbe, FMB, RoboSet and Spoc. 

\textbf{Vision language instruction data}: We use the datasets described in \Cref{sec:environments}. 

When sampling batches for any SFT training, we weight each dataset defined in \Cref{tab:datasets} equally, except OpenX is upsampled 2x, Procgen 2x, and all the VQA data is upsampled 3x.

%% file: supp/experiments.tex
\section{Evaluation Details}
\label{sec:eval-details} 

Overall, we use the standard evaluation settings per environment as defined by the prior work. We detail the evaluation settings for each environment below:

\textbf{MetaWorld}: We evaluate the generalization performance of our agent on Metaworld unseen starting states over 5 episodes each, totaling to 450 evaluation trajectories in total. We use the simulator's notion of success to define the success rate.

\textbf{Atari}: We evaluate the in-distribution performance of our model on 44 Atari games, by using the same setup as the DQN replay dataset, which in turn relies on the Dopamine~\citep{castro18dopamine} framework. Specifically, we use the $\texttt{\{GameID\}NoFramskip-v4}$ version of the ALE simulator~\citep{bellemare13arcade}. We conduct 10 rollouts over all 44 Atari games and average their respective human-normalized scores. The human-normalized scores follow the same protocol as~\citep{lee2022multi, reed2022generalist}:
$$
\text{score}_\text{normalized}(s)=\frac{|s-\text{score}_\text{random}|}{\text{score}_\text{human}-\text{score}_\text{random}}
$$
Through our experiments, we have found that GEA-Base performs better according to the human-normalized average score than GEA (40.3 vs 32.71). The result is not surprising, as we do not perform any Atari online RL finetuning on the GEA-Base model, and hence, performance can degrade at the expense of much better performance on Habitat Pick and Procgen. To solve this issue, one should co-train on all tasks that support fast online simulation. However, since Atari is structurally similar to Procgen, and Procgen supports procedural level generation to test generalization, we choose not to perform online RL finetuning on Atari.

\textbf{BabyAI}: We evaluate each task from \citet{babyai_iclr19} and evaluate over 100 random episodes for each of the 17 tasks, resulting in 1700 total episodes for the numbers reported in \Cref{table:task-gen}. Each episode has a different environment state and new language instruction. 

\textbf{Procgen}: We follow the test evaluation setting \citet{mediratta2023generalization} which uses the ``easy" mode setting of the game. We evaluate 50 episodes for each of the 16 games. Like Atari, the performance is evaluated in Procgen via the min-max normalized per-game scores from a random policy and the a PPO expert policy using the reported scores from \citet{cobbe2020leveraging}. 

\textbf{CALVIN}: We use the $ ABC \rightarrow D$ evaluation setting. This means that during our evaluation, the table background and the language instructions are unseen. We report results over the full 1k evaluation episodes defined by \citet{mees2022calvin}. 

\textbf{LangR}: We follow the standard evaluation settings from \citet{szot2021habitat}. Like the original work, our numbers are reported over the 9 unseen language instruction splits. We evaluate in 100 episodes for each of the 9 evaluation splits. Each test episode is also in an unseen house layout.

\textbf{Habitat Pick/Place/Nav}: We follow the evaluation split and settings from \citet{szot2021habitat} and evaluate the agent for 500 episodes in unseen house layouts. This version of the task where the agent has to operate from a language instruction of which object to pick is a harder version of the original geometric goal task and has been employed by prior works \cite{harish2025reinforcement,szot2024grounding}. 

\textbf{Maniskill}: We use the standard evaluation settings from \citet{gu2023maniskill2}. Aligning with the generated dataset for Maniskill, we evaluate in the ``StackCube", ``PegInsertionSide", ``PlugCharger", ``PushCube", and ``PickCube" tasks. Each of the 5 tasks are evaluated for 100 episodes each. 

\textbf{Android Control}: We evaluate on the full 1,540 test episodes from \citet{li2024effects}. We measure the per-step success rate, meaning the percent of the time the agent predicts the right action based on the ground truth test episode. This is distinct from the episode level success rate which requires the agent to predict every action in the trajectory correctly. We report the success rate under the more challenging setting of following high-level instructions only, without per-step low-level instructions. We consider an action as predicted correctly if it is within 5 of the horizontal and vertical patches defined in the dataset generation from \Cref{supp:dataset}. Any entered text is considered correct if the correct text is contained in the entered text or the entered text is contained in the correct text.

\begin{figure*}[t]
  \centering
  \begin{subfigure}[b]{0.35\textwidth}
    \centering
    \includegraphics[width=\textwidth]{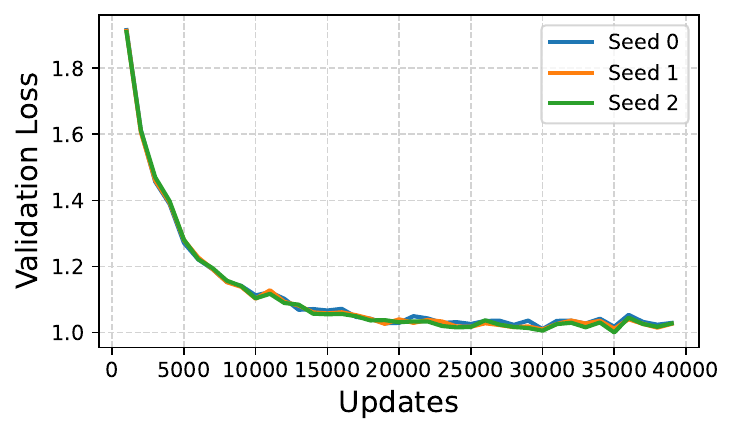}
    \caption{Validation Loss (\geabasesmall).}
    \label{fig:var-loss}
  \end{subfigure}
  \begin{subfigure}[b]{0.55\textwidth}
    \centering
    \resizebox{1.0\textwidth}{!}{
      \input{tables/seed_ablation}
    }
    \caption{Evaluation success rates (\geabasesmall).}
    \label{fig:var-eval}
  \end{subfigure}
  \caption{
    Variance in training jobs and evaluation for \gea. We train three different random seeds for \geabasesmall. The left shows the validation loss during SFT is similar between each random seed. The right shows the resulting online evaluation for these three random seeds across six of the benchmarks, along with the combined averages and standard deviation per benchmark. While the standard deviation is low, it is still a couple of percent on some benchmarks despite the validation loss being very similar.
  }
  \label{fig:seed_var}
\end{figure*}

\section{Further Experimental Details}

\subsection{Additional Baseline Details}
\label{sec:baselines} 

For each benchmark, to the best of our knowledge, we report the method from prior work with the highest performance. In this section, we add more details about these baselines from prior works in \Cref{table:task-gen} and any differences in the evaluation settings and method assumptions from \gea. We source methods from prior work that train a single policy over all tasks in a benchmark. We consider methods that only train on tasks from a single benchmark as ``specialist" and methods that train across multiple benchmarks as ``generalist". Note that this means we do not compare against methods that train policies on individual tasks from the benchmark. For example, in Procgen, we do not compare to the performance of \citet{cobbe2020leveraging} since a separate policy is trained for each of the 16 tasks. Instead, we only compare to methods that train a single policy across multiple tasks.

\textbf{Meta-World}: Many prior works report performance on the Meta-World benchmark~\cite{haldar2024baku}, but fewer works train and evaluate over the full set of 45 tasks. Gato~\cite{reed2022generalist} trains over all the Meta-World tasks and reports an average performance of $ 87.0\%$ success rate. \footnote{We reference the Gato numbers from Table 8 of the TMLR paper version: https://openreview.net/pdf?id=1ikK0kHjvj}
Unlike \gea, Gato also takes as input the proprioceptive state in Meta-World. \citet{reed2022generalist} also does not clarify if the Meta-World evaluation is performed over unseen state configurations or using the same states seen during training. \gea is evaluated on unseen state configurations. The \gea Meta-World numbers are reported in the same setting as \citet{szot2024grounding} with the same inputs.

\textbf{CALVIN:} To the best of our knowledge, there are no generalist agents that report the performance on CALVIN in addition to other benchmarks. RoboFlamingo~\cite{li2023vision} also adapts an MLLM for control by finetuning it with supervised learning. We include this specialist agent since it also leverages finetuning MLLMs, to demonstrate the gains from scaling to a generalist model with \gea. Compared to \gea, RoboFlamingo uses the gripper camera, image augmentations during training and a longer context length. 3D Diffuser Actor~\cite{ke20243d} is the state-of-the-art specialist system for CALVIN $ ABC \rightarrow D$ setting and narrowly outperforms \gea. However, as mentioned in the main text, 3D Diffuser Actor also assumes input to 3D pointclouds features. This method uses the head and gripper RGBD cameras to extra pointclouds of the scene. These pointclouds are then converted into a 3D feature cloud using a pretrained CLIP model. This method also simplifies the problem by compressing longer sequences of actions into end-effector keyposes.

\textbf{Maniskill:} We compare to the numbers using RGBD in Table 2 and 3 of \citet{gu2023maniskill2}. Unlike these numbers, \gea uses only RGB inputs and no depth images. These results train a single policy per-task which is technically narrower than our definition of ``Specialist Agent" which requires training one policy on all tasks from the benchmark. However, we still include these baselines to situate our Maniskill results. \gea outperforms doing imitation learning with the exact same demonstrations as from Table 2 of \citet{gu2023maniskill2}. The superior results of $ 47.8\%$ success are from Table 3 of \citet{gu2023maniskill2}, which train with DAPG~\cite{rajeswaran2017learning} and PPO. \gea does not train with RL in this task. \citet{hansen2023td} reports higher success rates in Maniskill2 tasks, but uses the ground truth state information instead of visual observations and trains a single policy per individual Maniskill2.

\textbf{Habitat Pick:} Other methods that report performance on the version of the Habitat Pick task that requires picking from the object name typically achieve low success rates~\cite{harish2025reinforcement,szot2024grounding,yenamandra2023homerobot}. We thus also compare to the expert policy that was used to generate the Habitat Pick training dataset as described in \Cref{supp:dataset}. Note that this expert policy was not trained in the evaluation scenes and thus also must generalize to unseen scenes. This expert policy has performance on par with pick skills trained in Habitat that operates from the much stronger assumption of a geometric goal input~\cite{szot2021habitat}. The specialist numbers from \cite{szot2024grounding} also finetune a MLLM with imitation learning and are evaluated in the exact same setting as \gea. 

\textbf{Habitat Place:} We follow the same evaluation criteria as Habitat Pick and compare to the expert policy trained with RL that was used to generate the expert demonstrations as described in \Cref{supp:dataset}.

\textbf{Procgen:} We compare against the BC test numbers from Figure 2 of \citet{mediratta2023generalization}, which use the same datasets as our setup. 
While Gato~\cite{reed2022generalist} also reports numbers in Procgen, we are not able to compare to these numbers because Gato reports performance relative to unknown score of the data collection policy. To the best of our knowledge the score of the data collection policy is not released. Thus, it is unclear how the Gato Procgen performance is normalized according to the standard Procgen normalization scores~\cite{cobbe2020leveraging} rendering a direct comparison impossible. 

\textbf{Atari:} For a generalist system, we compare with Gato~\cite{reed2022generalist} which as Table 8 shows of \citet{reed2022generalist} shows, achieves $ 30.9$ normalized score. Multi-Game Decision Transformers~\cite{lee2022multi} achieves $ 85$ normalized score in the same scoring setting. This method was trained with offline RL based on conditioning the training on the reward-to-go~\cite{chen2021decision}. 

\textbf{Habitat Nav:} We compare against the success rate of the navigation policy from \citet{szot2021habitat}. This policy is trained with RL on the same training set of episodes and evaluated on the same testing set of episodes as \gea. However, this policy operates on the geometric goal specification of the receptacle rather than the receptacle name. Additionally, this policy also takes an egomotion sensor as input, whereas \gea does not, simplifying the problem. 

\textbf{BabyAI:} We compare against Gato~\cite{reed2022generalist} which as Table 8 in \citet{reed2022generalist} shows, achieves $ 93.2$ normalized score. While \citet{reed2022generalist} does not clarify this detail, presumably the normalization is with respect to a perfect expert policy, so the normalized score is equal to the success rate. Gato trains with far more data than \gea with 4.61M episodes.

\textbf{AndroidControl:} We compare against using a Set-of-Mark prompting with GPT-4o. The Set-of-Mark was implemented using the Ferret-UI model~\cite{li2024ferret} for UI element detection to generate the marks with GPT-4o to determine the action. \gea and this baseline are evaluated under the same success criteria. We do not compare to the numbers from the original AndroidControl paper~\cite{li2024effects} since it uses different evaluation criteria from ours described in \Cref{sec:eval-details} and the code for the evaluation in \citet{li2024effects} is not publically released. 

\textbf{LangR:} We compare against the state-of-the-art numbers from \citet{szot2024grounding}. This method was trained with RL over the same set of training episodes as \gea.

\subsection{Ablation Analysis Setting}
\label{supp:analysis-dataset} 
\label{supp:analysis-method} 

For the analysis experiments, we used a reduced subset of the total dataset. Specifically, we use the Meta-World, CALVIN, Habitat Pick, BabyAI, Procgen, and AndroidControl datasets. Datasets from the other domains are excluded. 

When training methods in the analysis setting, we keep all the same settings as from the main \gea experiments with the hyperparameters described in \Cref{supp:sft}. However, we reduce the number of updates to 40k and use a global batch size of 256 across 2 nodes of 8 H100 GPUs each. All results in the analysis section are reported in the \lovsmall setting unless specified otherwise.

\section{Further Results}

\subsection{Benchmark Per-Task Success Rate}
\label{supp:breakdown} 
We breakdown the performance of the \gea model reported in \Cref{table:task-gen} per individual task for the benchmarks of Meta-World (\Cref{table:metaworld-break}), CALVIN (\Cref{table:calvin-break}), Procgen (\Cref{table:procgen-break}), Maniskill (\Cref{table:maniskill-break}), LangR (\Cref{table:langR-break}), and BabyAI (\Cref{table:babyai-break}).

\subsection{Habitat Pick RL Finetuning}
\label{supp:habpick-rl} 

\begin{figure}[h!] 
  \centering
  \includegraphics[width=1.0\columnwidth]{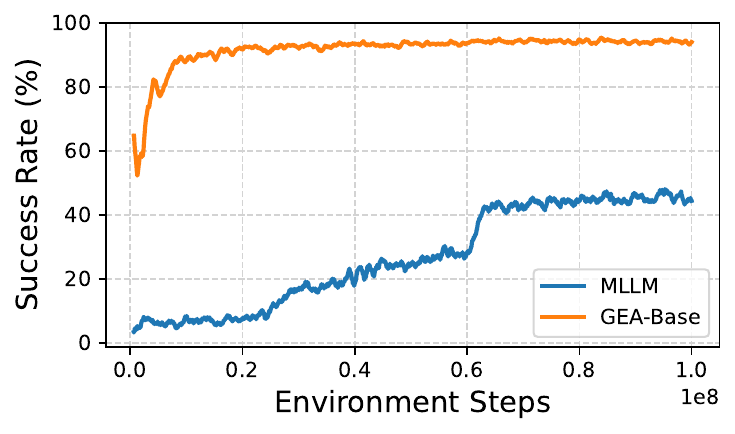}
  \caption{
    Success rate on the Habitat Pick task of \geabase and the base \mllm when finetuning with online RL. Displayed are success rates on the training dataset used in the RL process.
  }
  \label{fig:pick-learning}
\end{figure}

Complimenting the results demonstrating the value of online learning from \Cref{fig:habpick-online}, in this section, we compare the RL sample efficiency of \geabase versus the base \mllm. \Cref{fig:pick-learning} shows that doing RL from the \geabase model is far more sample efficient and converges to much higher performance than doing RL on the \mllm model. The \geabase model is trained with SFT on demonstrations from the Habitat Pick task, so it is expected that its performance will start higher. However, the \mllm model is never able to make up for the performance gap, even with continued RL finetuning.

\subsection{Model Variance Analysis}
\label{supp:model-var} 

In this section, we analyze the sensitivity of training \geasmallbase with different random seeds. Specifically, we vary the random seed used to initialize the model, all aspects of the algorithm, and dataset sampling. We then train the \geasmallbase model in the analysis setting from \Cref{supp:analysis-method}. The results in \Cref{fig:seed_var} demonstrate that while the training and validation curves are very similar. The online evaluation performance of \geasmallbase does have some variance per random seed within a couple of percentage points.

\subsection{Domain Transfer Analysis}
\label{supp:transfer} 

\begin{figure}[h!] 
  \centering
  \includegraphics[width=1.0\columnwidth]{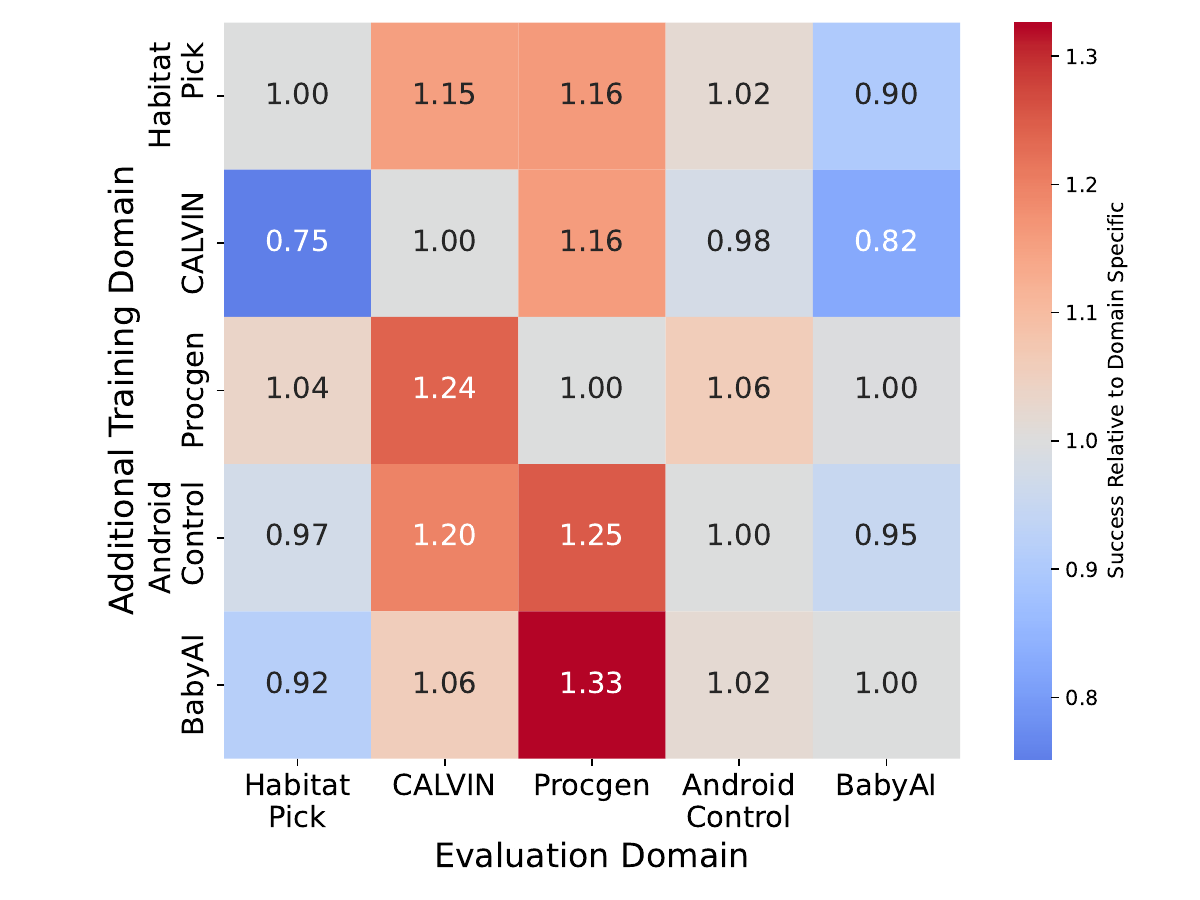}
  \caption{
    Each square represents the success rate of \geabasesmall trained with the datasets from the two domains indicated by the column and row name and evaluated on the domain indicated by the column. Success rates are scaled by training on only data from that domain, meaning each column is normalized relative to the diagonal. A more blue color means negative transfer and a more red color means positive transfer. 
  }
  \label{fig:data-pair-ablation}
\end{figure}

In \Cref{fig:data-pair-ablation} we study how performance transfers between domains by training the \geabasesmall model on every possible pair of datasets from BabyAI, CALVIN, Habitat Pick, Android Control and Procgen. We train the model in the same analysis setting as from \Cref{supp:analysis-dataset}. We compare the performance of the dataset pair to only training on one of the datasets (the same as the ``Domain Specific" results from \Cref{table:llm-pretraining}). The results in \Cref{fig:data-pair-ablation}, with more blue colors for negative transfer and more red colors for positive transfer, show that some domains such as Procgen and CALVIN enormously benefit from data in other domains. Android Control slightly benefits from Procgen, another discrete control task. On the other hand, Habitat Pick and BabyAI have negative transfer from a variety of tasks. Despite this negative transfer, \Cref{table:llm-pretraining} still demonstrates that training on all the data across all these domains improves the performance.

\begin{table}[h!] 
  \centering
  \input{tables/metaworld_breakdown}
  \caption{
    Meta-World per-task success rate breakdown.
  }
  \label{table:metaworld-break} 
\end{table}

\begin{table}[h!] 
  \centering
  \input{tables/calvin_breakdown}
  \caption{
    CALVIN per-task success rate breakdown. Note the tasks are not equally represented in the evaluation episodes. 
  }
  \label{table:calvin-break} 
\end{table}

\begin{table}[h!] 
  \centering
  \input{tables/procgen_breakdown}
  \caption{
    Procgen per-game score breakdown. 
  }
  \label{table:procgen-break} 
\end{table}

\begin{table}[h!] 
  \centering
  \input{tables/maniskill_breakdown}
  \caption{
    Maniskill per-task score breakdown. 
  }
  \label{table:maniskill-break} 
\end{table}

\begin{table}[h!] 
  \centering
  \input{tables/langR_breakdown}
  \caption{
    LangR per-task score breakdown. 
  }
  \label{table:langR-break} 
\end{table}

\begin{table}[h!] 
  \centering
  \input{tables/babyai_breakdown}
  \caption{
    BabyAI per-task score breakdown. 
  }
  \label{table:babyai-break} 
\end{table}

\begin{table}[h!] 
  \centering
  \input{tables/atari_breakdown}
  \caption{
    Atari success rate breakdown.
  }
  \label{table:atari-break} 
\end{table}

%% file: tables/seed_ablation.tex
\begin{tabular}{P{0.14\columnwidth}P{0.14\columnwidth}P{0.14\columnwidth}P{0.14\columnwidth}P{0.14\columnwidth}P{0.14\columnwidth}P{0.14\columnwidth}}
\toprule
 & \textbf{HabPick} & \textbf{Procgen} & \textbf{CALVIN} & \textbf{BabyAI} & \textbf{Meta-World} & \textbf{Android Control} \\
\midrule
Seed 0 &  46.5  & \textbf{  29.0  } & \textbf{  44.5  } & \textbf{  82.6  } &  82.7  &  48.4  \\
Seed 1 &  49.0  &  25.4  &  42.5  &  80.0  &  86.7  & \textbf{  50.1  } \\
Seed 2 & \textbf{  53.0  } &  27.3  &  44.5  &  82.4  & \textbf{  88.4  } &  49.2  \\
Combined &  49.5 {\scriptsize $ \pm$ 3.3  } &  27.2 {\scriptsize $ \pm$ 1.8  } &  43.8 {\scriptsize $ \pm$ 1.2  } &  81.7 {\scriptsize $ \pm$ 1.5  } &  85.9 {\scriptsize $ \pm$ 3.0  } &  49.2 {\scriptsize $ \pm$ 0.8  } \\

\end{tabular}

%% file: tables/metaworld_breakdown.tex
\begin{tabular}{P{0.50\columnwidth}P{0.50\columnwidth}}
\toprule
 & \textbf{\gea} \\
\midrule
assembly &  86.0  \\
basketball &  100.0  \\
button-press-topdown &  100.0  \\
button-press-topdown-wall &  100.0  \\
button-press &  100.0  \\
button-press-wall &  92.0  \\
coffee-button &  100.0  \\
coffee-pull &  100.0  \\
coffee-push &  100.0  \\
dial-turn &  100.0  \\
disassemble &  88.0  \\
door-close &  100.0  \\
door-open &  100.0  \\
drawer-close &  100.0  \\
drawer-open &  100.0  \\
faucet-open &  100.0  \\
faucet-close &  100.0  \\
hammer &  100.0  \\
handle-press-side &  100.0  \\
handle-press &  100.0  \\
handle-pull-side &  80.0  \\
handle-pull &  100.0  \\
lever-pull &  80.0  \\
peg-insert-side &  100.0  \\
peg-unplug-side &  98.0  \\
pick-out-of-hole &  60.0  \\
pick-place &  92.0  \\
pick-place-wall &  82.0  \\
plate-slide &  100.0  \\
plate-slide-side &  96.0  \\
plate-slide-back &  100.0  \\
plate-slide-back-side &  100.0  \\
push-back &  90.0  \\
push &  100.0  \\
push-wall &  100.0  \\
reach &  60.0  \\
reach-wall &  94.0  \\
shelf-place &  98.0  \\
soccer &  76.0  \\
stick-push &  100.0  \\
stick-pull &  98.0  \\
sweep-into &  90.0  \\
sweep &  100.0  \\
window-open &  100.0  \\
window-close &  100.0  \\

\end{tabular}

%% file: tables/calvin_breakdown.tex
\begin{tabular}{P{0.50\columnwidth}P{0.50\columnwidth}}
\toprule
 & \textbf{\gea} \\
\midrule
turn off led &  87.0  \\
move slider left &  100.0  \\
rotate red block right &  69.0  \\
open drawer &  100.0  \\
rotate red block left &  95.5  \\
push pink block right &  100.0  \\
push blue block right &  65.2  \\
push red block left &  85.7  \\
push pink block left &  87.1  \\
push red block right &  31.0  \\
push blue block left &  88.6  \\
push into drawer &  86.7  \\
rotate pink block left &  100.0  \\
turn on lightbulb &  97.6  \\
rotate pink block right &  93.5  \\
rotate blue block right &  78.6  \\
turn off lightbulb &  97.1  \\
lift blue block table &  100.0  \\
close drawer &  100.0  \\
rotate blue block left &  95.8  \\
move slider right &  95.4  \\
turn on led &  96.4  \\
lift blue block slider &  63.3  \\
lift pink block table &  100.0  \\
lift red block slider &  73.1  \\
lift red block table &  83.3  \\
lift pink block slider &  96.0  \\

\end{tabular}

%% file: tables/procgen_breakdown.tex
\begin{tabular}{P{0.50\columnwidth}P{0.50\columnwidth}}
\toprule
 & \textbf{\gea} \\
\midrule
bigfish &  43.1  \\
bossfight &  54.2  \\
caveflyer &  27.7  \\
chaser &  54.0  \\
coinrun &  66.0  \\
dodgeball &  3.4  \\
fruitbot &  84.8  \\
heist &  32.0  \\
leaper &  26.0  \\
maze &  58.0  \\
miner &  27.0  \\
climber &  46.2  \\
ninja &  62.0  \\
plunder &  4.5  \\
jumper &  50.0  \\

\end{tabular}

%% file: tables/maniskill_breakdown.tex
\begin{tabular}{P{0.50\columnwidth}P{0.50\columnwidth}}
\toprule
 & \textbf{\gea} \\
\midrule
StackCube &  4.0  \\
PegInsertionSide &  0.0  \\
PlugCharger &  0.0  \\
PushCube &  52.0  \\
PickCube &  12.0  \\

\end{tabular}

%% file: tables/langR_breakdown.tex
\begin{tabular}{P{0.50\columnwidth}P{0.50\columnwidth}}
\toprule
 & \textbf{\gea} \\
\midrule
rephrasing &  84.0  \\
referring expressions &  16.0  \\
spatial relationships &  0.0  \\
context &  34.0  \\
irrelevant text &  86.0  \\
multiple rearrangements &  82.0  \\
novel objects &  96.0  \\
multiple objects &  0.0  \\
conditional instructions &  52.0  \\

\end{tabular}

%% file: tables/babyai_breakdown.tex
\begin{tabular}{P{0.50\columnwidth}P{0.50\columnwidth}}
\toprule
 & \textbf{\gea} \\
\midrule
GoToRedBallGrey &  88.0  \\
GoToRedBall &  97.0  \\
GoToRedBallNoDists &  100.0  \\
GoToObj &  100.0  \\
GoToObjS4 &  100.0  \\
GoToLocalS8N7 &  93.0  \\
GoToRedBlueBall &  92.0  \\
GoToDoor &  100.0  \\
Open &  76.0  \\
OpenRedDoor &  100.0  \\
OpenDoorColor &  100.0  \\
OpenDoorLoc &  80.0  \\
OpenTwoDoors &  100.0  \\
OpenRedBlueDoors &  100.0  \\
Pickup &  41.0  \\
PickupLoc &  88.0  \\
PickupDist &  94.0  \\

\end{tabular}

%% file: tables/atari_breakdown.tex
\begin{tabular}{P{0.50\columnwidth}P{0.50\columnwidth}}
\toprule
 &  \textbf{\gea} \\
\midrule
Alien & 6.1  \\
Amidar & 0.0  \\
Assault & 86.5  \\
Asterix & 2.3  \\
Atlantis & 0.0  \\
BankHeist & 8.9  \\
BattleZone & 39.2  \\
BeamRider & 3.1  \\
Boxing & 0.0  \\
Breakout & 0.0  \\
Centipede & 50.0  \\
ChopperCommand & 30.2  \\
CrazyClimber & 0.0  \\
DemonAttack & 20.8  \\
DoubleDunk & 300.0  \\
Enduro & 0.0  \\
FishingDerby & 0.0  \\
Freeway & 81.1  \\
Frostbite & 2.7  \\
Gopher & 0.0  \\
Gravitar & 0.0  \\
Hero & 0.0  \\
IceHockey & 0.0  \\
Jamesbond & 0.0  \\
Kangaroo & 38.5  \\
Krull & 373.0  \\
KungFuMaster & 0.0  \\
MsPacman & 27.9  \\
NameThisGame & 0.0  \\
Phoenix & 0.0  \\
Pong & 0.0  \\
Qbert & 0.6  \\
Riverraid & 0.0  \\
RoadRunner & 33.0  \\
Robotank & 307.2  \\
Seaquest & 0.3  \\
SpaceInvaders & 52.7  \\
StarGunner & 1.4  \\
TimePilot & 0.0  \\
UpNDown & 0.0  \\
VideoPinball & 0.0  \\
WizardOfWor & 0.0  \\
YarsRevenge & 0.0  \\
Zaxxon & 0.0  \\

\end{tabular}